\documentclass[journal]{IEEEtran}

\ifCLASSINFOpdf
  \usepackage[pdftex]{graphicx}
\else
  \usepackage[dvips]{graphicx}
\fi

\ifCLASSOPTIONcompsoc
  \usepackage[caption=false,font=normalsize,labelfont=sf,textfont=sf]{subfig}
\else
  \usepackage[caption=false,font=footnotesize]{subfig}
\fi

\usepackage{amsmath,amsfonts,amsthm,amssymb}

\usepackage{url}
\usepackage{multirow}
\usepackage{pdflscape}
\usepackage{listings}
\usepackage{caption}
\usepackage{hyperref}

\begin{document}

\title{StarAlgo: A Squad Movement Planning\\ Library for StarCraft using\\ Monte Carlo Tree Search and Negamax}

\author{Mykyta Viazovskyi, Michal~\v{C}ertick\'{y}
\thanks{Mykyta Viazovskyi is with Department of Software Engineering at Faculty of Information Technologies, Czech Technical University in Prague. Email: {\scriptsize \tt{viazomyk@fit.cvut.cz}}}
\thanks{Michal~\v{C}ertick\'{y} is with Artificial Intelligence Center, Department of Computer Science, Czech Technical University in Prague. Email:~\tt{\scriptsize certicky@agents.fel.cvut.cz}}%
}

%

\maketitle
\begin{abstract}
Real-Time Strategy (RTS) games have recently become a popular testbed for artificial intelligence research. They represent a complex adversarial domain providing a number of interesting AI challenges.
There exists a wide variety of research-supporting software tools, libraries and frameworks for one RTS game in particular -- StarCraft: Brood War. These tools are designed to address various specific sub-problems, such as resource allocation or opponent modelling so that researchers can focus exclusively on the tasks relevant to them.
We present one such tool -- a library called StarAlgo that produces plans for the coordinated movement of squads (groups of combat units) within the game world. StarAlgo library can solve the squad movement planning problem using one of two algorithms: Monte Carlo Tree Search Considering Durations (MCTSCD) and a slightly modified version of Negamax. We evaluate both the algorithms, compare them, and demonstrate their usage. 
The library is implemented as a static C++ library that can be easily plugged into most StarCraft AI bots. 
\end{abstract}


%
\IEEEpeerreviewmaketitle


\section{Introduction}\label{secIntro}

Games have traditionally been used as domains for Artificial Intelligence (AI) research since they represent well-defined challenges with varying degrees of complexity. They are also easy to understand and provide a way to compare the performance of AI algorithms and that of human players~\cite{maly2018multi}. 
After recent success with popular board games like Go \cite{silver2016mastering}, the attention of researchers has turned to a more complex challenge represented by Real-Time Strategy (RTS) games. 

RTS is a genre of video games in which players manage economic and strategic tasks by gathering resources and building bases, increase their military power by researching new technologies and training units, and lead them into battle against their opponent(s)~\cite{certicky2017current}.
RTS games are played in real time, meaning that the actions must be decided in fractions of a second and hundreds of player-issued actions are being executed at any given time \cite{buro2012real}. The domain itself is partially observable and non-deterministic. 
From the perspective of AI research, RTS games pose interesting challenges in the areas of planning, dealing with uncertainty, domain knowledge exploitation, task decomposition, spatial reasoning, and machine learning \cite{Survey2013, churchill2016starcraft}.


The high complexity of the RTS game domain encourages its decomposition into smaller sub-problems, such as unit micromanagement in combat, threat-aware pathfinding, resource allocation, opponent modelling or build order optimization. However, many of these sub-problems cannot be isolated from the others without oversimplifying -- in order to effectively solve one problem, other problems often need to be considered too. For example, in order to solve the micromanagement of combat units, one might need a threat-aware pathfinding algorithm (to be able to surround the enemy units without getting killed) or opponent modelling algorithm to predict how the opponent will react in specific combat situations.

It is a common practice for AI researchers to focus only on one specific RTS sub-problem at any given time and use existing third-party tools as building blocks to solve the others. Various ready-to-use software tools, designed to solve different RTS sub-problems, are already available for one of the most popular RTS games~--~StarCraft: Brood War, released in 1998 by Blizzard Entertainment. For example, Brood War Terrain Analyzer (BWTA and BWTA2~\cite{bwta2}) and Brood War Easy Map (BWEM~\cite{bwem}) are two widely used libraries able to analyze the map and return key regions, chokepoints, and base locations. BWAPI Standard Add-on Library (BWSAL~\cite{bwsal}) is a collection of ready-to-use solutions for different sub-tasks, including worker management, scouting, research, base building, etc. A widely-used library called SparCraft~\cite{ualberta} provides a combat simulator that can be used to predict the battle outcomes and the Build Order Search System (BOSS) can be used to search for optimal base construction plans. TorchCraft~\cite{torchcraft} is another interesting tool, designed to simplify the application of machine learning algorithms to StarCraft.

In this work, we focus on the problem of controlling the movement of squads (groups of combat units) within the game world. This problem was already tackled by Alberto Uriarte and Santiago Onta\~{n}\'{o}n~\cite{uriarte14a} in 2014. They demonstrated how a specific variant of Monte Carlo Tree Search algorithm (MCTSCD), together with combat simulation, can be used to produce reasonable squad movement plans for the following few minutes. Unfortunately, their solution was implemented only as a prototype crudely integrated into their own bot for demonstration purposes, and could not be easily reused by other researchers and bot programmers.    


In order to give the research community an easy-to-use tool for the squad movement planning problem, we present the library called StarAlgo. It implements the MCTSCD algorithm, as described by Uriarte and Onta\~{n}\'{o}n, as well as a modified Negamax algorithm. The library provides a set of functions and classes that allow the users to find the most effective way to control their squads (attacking, retreating, defending) while taking into account the compositions and locations of friendly and enemy squads, map layout, chokepoints, base locations, etc. The library uses BWTA2 for map analysis and SparCraft for combat simulation. 

It is freely available as an open-source project at \url{https://github.com/Games-and-Simulations/StarAlgo}.

\section{Library Design}

\subsection{Library Format}
\label{library_format}

A C++ library for BWAPI-powered StarCraft AI bots can be either statically or dynamically linked. We decided to distribute the library statically linked for the following pragmatic reasons:
\begin{itemize}
    \item it is compiled directly into the bot executable
    \item it saves execution latency
    \item all functionality is guaranteed to be up to date (no versioning problems)
\end{itemize}

The BWAPI interface is a shared library itself, which might add some delay on its own. With computationally intensive problems in the context of StarCraft AI, it is always helpful to optimize the performance. The rules of StarCraft AI tournaments, such as SSCAIT, AIIDE and CIG~\cite{tournaments2018}, enforce maximum time allowed for the bot's computational tasks.   

The size of the statically linked library is approximately 25 MB, which is sufficiently low.  

Since the library provides a solution to a specific well-defined subproblem of squad management, it should be easy to integrate it to many current bots. These often have some kind of a modular structure -- for example, UAlbertaBot\footnote{\url{https://github.com/davechurchill/ualbertabot}} by D.~Churchill has a ``CombatManager'' module, which takes care of unit squad movement. This is a good place to use the StarAlgo library. We provide an example integration of StarAlgo library to UAlbertaBot at \url{https://github.com/Games-and-Simulations/StarAlgo/tree/master/examples/}.


\subsection{Library Architecture}
\label{lib_arch}

Overall, the project consists of 23 classes. The most crucial ones are: \textit{AbstractGroup, ActionGenerator, CombatSimulator, EvaluationFunction, GameNode, GameState, MCTSCD, RegionManager} and \textit{UnitInfoStatic} (see Figure \ref{class_diagram}). These make up the absolute core of the library, and we should examine each of them.

\begin{minipage}{\linewidth}
\begin{center}
\includegraphics[width=\textwidth]{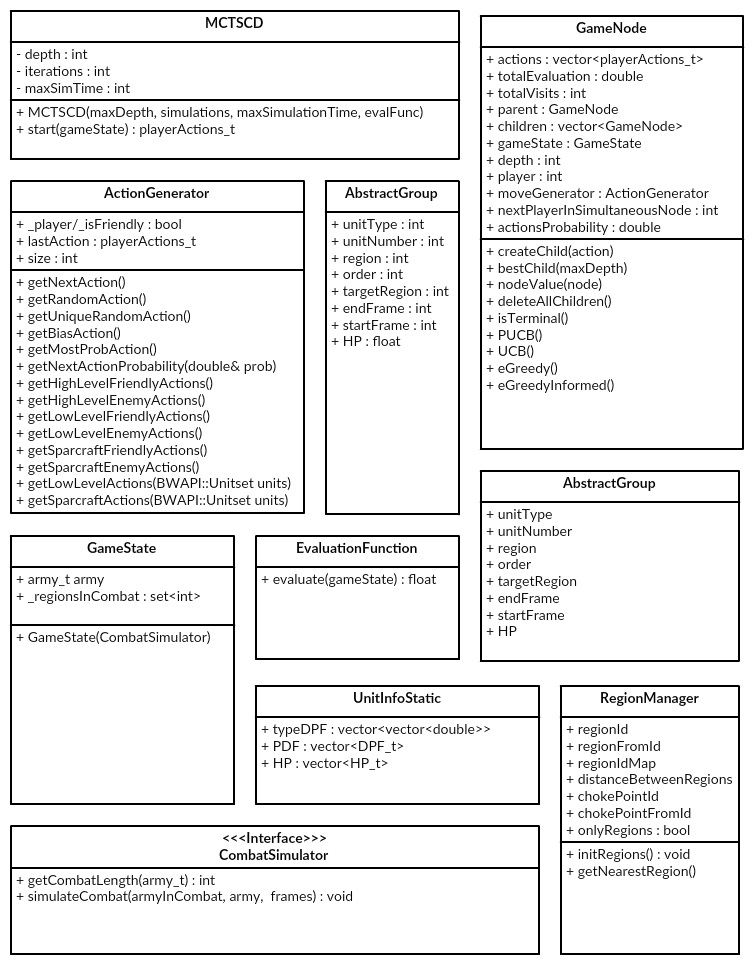}
\captionof{figure}{The most important StarAlgo classes.\label{class_diagram}}
\end{center}
\end{minipage}
\vspace{.8cm}

\textbf{AbstractGroup}. This class represents a unit squad - a small group of units located in the same region on the map. Actual division of the units into the squads is up to each bot. The AbstractGroup class keeps track of various information related to the squad.

\textbf{ActionGenerator} is responsible for generating all the valid actions for all the player's squads. The actions serve as operators for the search.

\begin{figure}
\begin{center}
\includegraphics[width=\columnwidth]{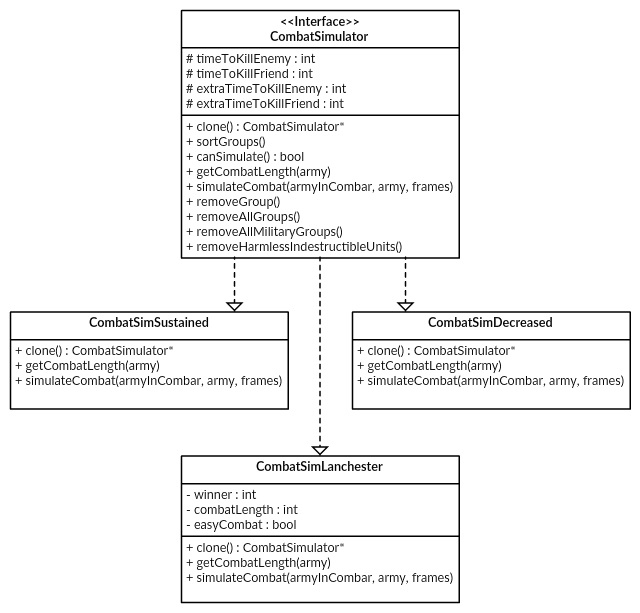}
\captionof{figure}{Combat simulation design class model\label{sim_des}}
\end{center}
\end{figure}

\textbf{CombatSimulator}. Whenever two opposing squads collide during the search (they meet in the same region), the outcome is resolved using a simplistic combat simulation~(see Figure \ref{sim_des}). This simulation can potentially be replaced by a more sophisticated one, such as SparCraft.  

\textbf{EvaluationFunction} is a simple, yet important class used during the search to evaluate the states. In our implementation, it counts the number of units of each type and multiplies it by their ``destroy score'' (the number of score points awarded to a player for killing the unit the end of the match). This can easily be replaced by a more sophisticated custom function. 

\textbf{GameState} represents one specific state of the game. It is defined by the squads of both players, their locations, actions being executed, and related temporal properties (durations and start times). The game state contains all the information required to generate  subsequent possible game states. 

\textbf{GameNode} is the core part of the search. It represents a single tree node and holds various search-related information, such as: \textit{actions} -- what actions are to be executed at this node, \textit{totalEvaluation} -- a numerical representation of how advantageous the actions are, \textit{totalVisits} -- how many times the node has been visited during the search, \textit{gameState} -- what is the actual situation on the map when the node is visited, and \textit{actionsProbability} -- how likely it is to have the selected actions actually appear in the game. 

\textbf{MCTSCD} is the implementation of the search itself. The search instance is defined by its depth, a number of iterations and maximum simulation time. All these values need to be provided by the user. The parameters are discussed in Section~\ref{library_use}.

\textbf{RegionManager} represents the map regions, chokepoints and all the information relevant to them. It allows all the other modules to instantly access the map information.


\textbf{UnitInfoStatic} provides useful information about all the available units in the game, such as their damage output, ability to attack air or ground units or their hit points.

\subsection{Algorithm Structure}

\begin{figure*}
\includegraphics[width=1\textwidth]{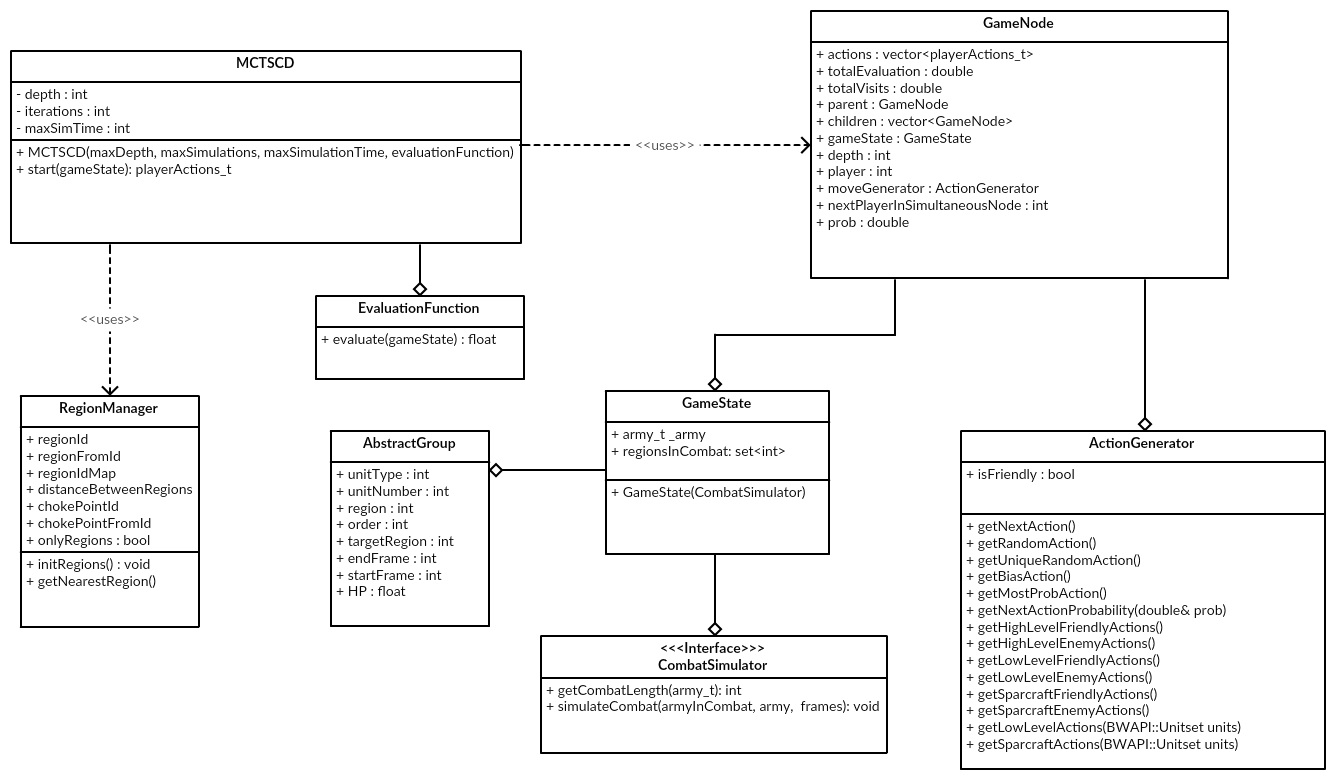}
\captionof{figure}{Class diagram of the MCTS search.\label{MCTS}}
\end{figure*}

The search algorithm~\cite{uriarte14a} is depicted on Figure \ref{MCTS}. The core of the algorithm is the MCTSCD class. It performs the search based on the possible actions in the GameNode and their evaluations produced by the EvaluationFunction.

The basic element of the search tree is the GameNode. Every GameNode contains the squad actions, the corresponding GameState and its evaluation and the number of visits of this node during the search. 

The GameState class is responsible for a more detailed representation of the game in terms of units and regions. GameState uses the CombatSimulator interface to approximate the outcomes of situations (states) when opposing squads meet in the same region. 


\section{Implementation}


We represent the game map as a graph, where each BWTA2 region corresponds to a single graph node and any two neighbouring regions are connected by an edge if they are accessible by land. The squad movement reasoning happens across this graph. 
For example, the map from Figure~\ref{graph_fig} would have 20 regions, 5 of which are not connected by land, which gives us a graph with 15 nodes, and 5 separate ones. Those nodes are only accessible by air units.

\vspace{.4cm}

\centerline{
\begin{minipage}{\linewidth}
\begin{center}
\includegraphics[width=0.9\textwidth]{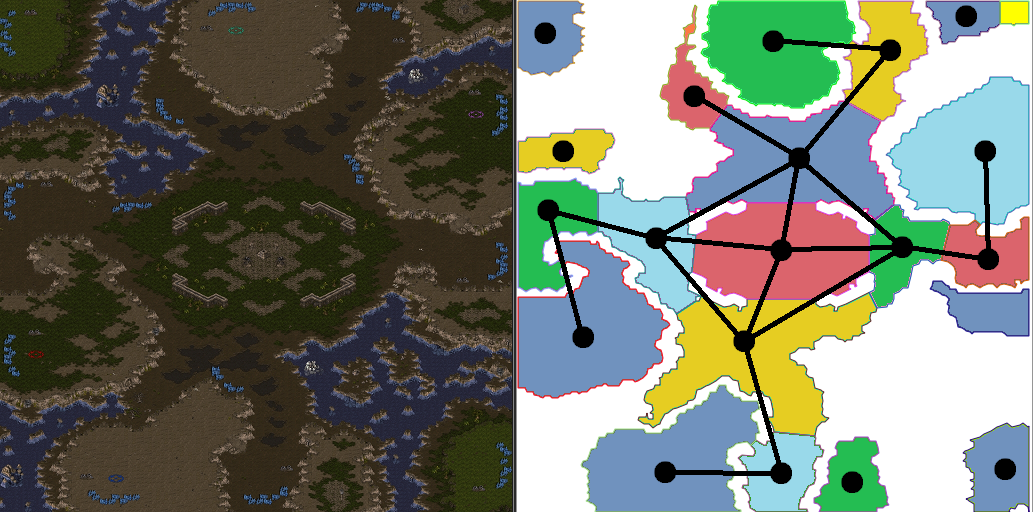}
\captionof{figure}{The game map represented as a graph.\label{graph_fig}}
\end{center}
\end{minipage}
}

\vspace{.4cm}

In order to prevail in the RTS game like StarCraft, we need to search within the vast state space of all the relevant actions. This includes not only the building placement and unit recruiting but also the movement of all the units. In StarCraft, there is an upper limit of 200 units controlled consecutively by a player (400 for Zerg race).
The high number of units to control and actions that each of them can execute prevent us from performing a complete search. Some kind of abstraction needs to be introduced.   
In order to decrease the search space, we only reason about the actions of unit squads instead of individual units, we decrease the spatial granularity to the region level instead of pixel precision and limit the search tree depth to represent only a few minutes of the game.

At the start of the search, all the combat units are used to define the initial game state. They are divided into squads~(AbstractGroups) and mapped to the game map graph. Next, the search with the simulations is performed starting with the initial game node and unfolding the search tree as the search proceeds. Finally, the sequence of actions~(plan) is returned. It needs to be sent back to the corresponding squads, which then start executing it. 




The StarAlgo library provides two algorithms that can be used for squad movement planning. The main one is MCTSCD, which seem to yield better results according to our experiments. The secondary algorithm is a modification of Negamax, which serves as a baseline.  

\subsection{Secondary Planning Algorithm: Negamax}
 
We provide an implementation of a slightly modified Negamax algorithm. Negamax search is a variant of MinMax search algorithm that relies on the zero-sum property of certain two-player games. 

In this case, the heuristic value of the search tree node is evaluated as the number of our units versus the number of enemy units in regions. This leads to a simple heuristic strategy of arranging our army in such a way that we have the military dominance at regions. This way, we are usually able to win the local battles. 

The goal of Negamax search is to find the node with the highest value, starting at the root node, representing an initial game state. The pseudocode below describes the basic Negamax base algorithm~\cite{breuker}, where we could limit the maximum search depth

The root node inherits its score from one of its children. The child node that ultimately sets the root node's best score also represents the best move to play~\cite{breuker}. Although the Negamax function only returns the node's best score as bestValue, our implementation  keeps both the evaluation and the node that holds the game state. 
In our alteration of Negamax, we omit the playerColor parameter. The heuristic evaluation function returns the values from the point of view of both players, since it considers the size of the army of each player.

\lstset{
xleftmargin=.1\columnwidth,
  numbers=left,
  mathescape,
  literate={∞}{$\infty$}{2}
  {s0}{$s_0$}{2}
  {n0}{$n_0$}{2}
  {n1}{$n_1$}{2}
}

\renewcommand{\lstlistingname}{Code}
{
\small
\begin{lstlisting}[caption=The pseudocode of the basic Negamax search algorithm.\vspace{.2cm}]
function negamax(node, depth, playerColor)
  if depth = 0 or node is terminal:
    return playerColor * value of node
 
  bestValue := -∞
  foreach child of node:
    v := -negamax(child, depth-1, -plColor)
    bestValue := max(bestValue, v)
  return bestValue
\end{lstlisting}
}
\hfill

\vspace{.3cm}
\subsection{Primary Planning Algorithm: Monte Carlo Tree Search Considering Durations (MCTSCD)} 

The idea behind this approach (as well as Monte Carlo methods in general) is to continuously sample random elements in order to obtain the results. It uses randomness to address a complex deterministic problem. 

In general, Monte Carlo methods have the following structure:
\begin{itemize}
    \item Define a domain of possible inputs.
    \item Generate inputs randomly from a probability distribution over the domain.
    \item Perform a deterministic computation on the inputs.
    \item Aggregate the results.
\end{itemize}

The Monte Carlo Tree Search is based on the Monte Carlo principle but still builds a search tree. Instead of just running random simulations from the current node, it uses the results of the simulations to compare the states and propagate the search recursively through the search tree.

Unlike Negamax, MCTS is able to deal with the high branching factor. It simulates possible game state progressions up to some predefined point in time. The key is to balance between the exploration and exploitation of the tree. In these terms, exploration is looking into undiscovered subtrees, and exploitation is expanding the most promising nodes. There is a variety of policies to simulate the game until the logical stop, the default one being the uniform random actions of the player~\cite{uriarte14a}.

The MCTS can be stopped at any moment during the search and the best solution found up to that point can be retrieved. 
The ability to stop the search at any time is the biggest practical difference between our Negamax and MCTS implementations. To control the computation time of Negamax, we must manually adjust the depth of the search to fit the time limits.

Another advantage of MCTS is its use of heuristic selection to explore the search tree. It does not try to unroll all the possible states. The search sub-trees following some highly undesirable states can potentially be abandoned. This is an essential part of what makes the algorithm effective.

\lstset{
xleftmargin=.1\columnwidth,
  numbers=left,
  mathescape,
  literate={∞}{$\infty$}{2}
  {s0}{$s_0$}{2}
  {n0}{$n_0$}{2}
  {n1}{$n_1$}{2}
}

\renewcommand{\lstlistingname}{Code}

{
\small
\begin{lstlisting}[caption=The pseudocode of MCTS Considering Durations.]
function MCTS(s_0)
  n_0 := CreateNode(s_0, 0)
  while withing computational budget:
    nl := TreePolicy(n_0)
    dp := DefaultPolicy(nl)
    BACKUP(nl, dp)
  return (BestChild(n_0)).action

function CreateNode(s, n_0)
  n.parent := n_0
  n.lastSimult := n_0.lastSimult
  n.player := PlayerToMove(s, n.lastSimult)
  if BothCanMove(s):
    n.lastSimult := n.player
  return n

function DefaultPolicy(n)
  lastSimult := n.lastSimult
  s := n.s
  while withing computational budget:
    p := PlayerToMove(s,lastSimult)
    if BothCanMove(s):
      lastSimult := p
    simulate game s with a 
    given policy and player
  return s.reward
\end{lstlisting}
}

\subsection{Using the Library}
\label{library_use}

To start with the library application, one needs to import the header files and instantiate the required classes.

Several parameters can be set for the MCTS search -- mainly the depth, number of iterations and maximum simulation time. The depth limits how deep the search should go down the tree. Upon reaching this limit, the node will be considered terminal. The number of iterations tells us how many children will be generated for each node. 
These arguments should be provided to the MCTSCD upon initialization. 

The search object needs to be created once and then invoked every time the bot needs a new plan. It is up to the bot creator, whether the algorithm should produce a new plan on every frame, or less often. For example, it is possible to run a very long search and rely on its result for a long time, as it would make predictions further to the future. On the other hand, it is possible to run short searches more often. The first approach might need some computation rebalancing on slow machines.

Probably the best way to use the library is to plan the squad movement for the following few minutes (run the search in a separate thread so the bot does not freeze the game) and then replan when the planned period nears its end or whenever the current plan gets inconsistent with the actual game state (due to unpredicted outcomes).  

\section{Experiments}

We tested the library by integrating it with the UAlbertaBot. This particular bot has an architecture that is easy to extend and makes the integration simple. 

\subsection{Case Study}

We investigated the structure of both the UAlbertaBot and the search and how those two could be merged with the least amount of effort for the end user.
Since our search does not use the same squads as UAlbertaBot, it was necessary to bind the squads to our own unit groups.

\begin{figure}
\caption{CombatManager of UAlbertaBot.}\label{combat_manager}
  \centering
\includegraphics[width=1\columnwidth]{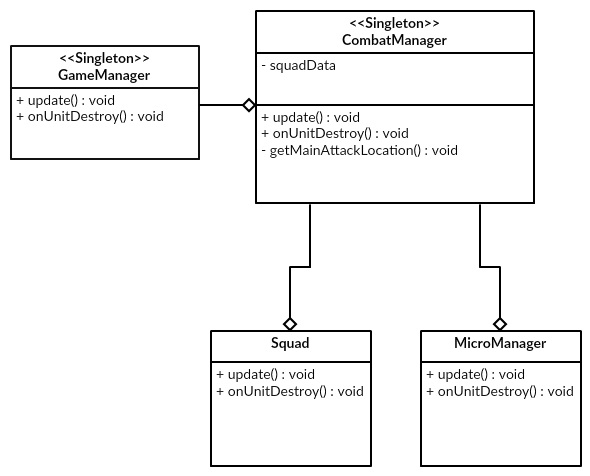}
\end{figure}

The CombatManager module (see Figure \ref{combat_manager}) of the UAlbertaBot was the correct place to start with the search integration. The execution of modules is hierarchical: the BWAPI library calls the \textit{onFrame} function of UAlbertaBot, which gets propagated to GameManager and then to CombatManager. The manager has access to the \textit{squadData}, a set of all squads of the player. This gives the control over the unit distribution across groups, which is exactly what is needed for the search.

The division of units into groups for the search is based on unit type and the region. The closer the squad formation is to the group, the more precise army coordination is achieved. Given a unit set, the search maps it to its internal structure, but the user has to keep track of the initial unit set since the actions of the resulting plan will need to be assigned back to it. The more the squad's units conform to the unit type and location of internal search groups, the better the search results are. Thus, we decided to split the army into several sets, based on their location. When the result is computed, it is assigned back to the squads.


The \textit{updateAttackSquads} method is the proper place to embody the search results. The result of each algorithm is the vector of locations corresponding to the vector of squads that the player currently owns.

\subsection{Comparison}

At the time of writing this text, our current implementation of MCTSCD is not able to deal with the partial observability of the game (this will be improved in the future). Therefore, we enabled the full game observability via the BWAPI setting, so both bots would see each other from the very beginning, and would not need to scout the map.


Both the Negamax and MCTSCD algorithms were integrated into UAlbertaBot. The bots played against the StarCraft in-game AI opponents. 


In all matches, the bots played the same race (Zerg) and the same build order (they built the same buildings and recruited the same units). This was done in order to minimize the randomness in the experiment.
The in-game AI was set to play the Protoss race. It selects among a few stable build orders.

We ran 160 games to show the comparative performance of MCTSCD versus Negamax.
Based on the winning data, the win ratio for MCTSCD was exactly $50\%$, while it was only $23.7\%$ for Negamax. 

\section{Conclusion}

We implemented a library for squad movement planning in StarCraft and released it as open-source project in hopes to help the RTS AI research community. It is distributed as a static C++ library and should be easy to integrate into most current BWAPI-based bots. The example integration into UAlbertaBot is included as a part of the project's Github repository.

Two planning algorithms are currently supported: Monte Carlo Tree Search Considering Durations (MCTSCD) and a modified Negamax. We compared both algorithms experimentally and discovered that MCTSCD performs considerably better.

At the time of writing, the library is in the early stage. We hope that the StarCraft AI community will contribute to its development in order to make it even more useful to them. The improvements planned for the future include adding support for partial observability, support for mixed unit type squads, increasing the level of spatial granularity, etc.


\bibliographystyle{IEEEtran}
\bibliography{references}{}

\end{document}